\documentclass[letterpaper]{article} 
\usepackage[submission]{aaai25}  
\usepackage{times}  
\usepackage{helvet}  
\usepackage{courier}  
\usepackage[hyphens]{url}  
\usepackage{graphicx} 
\usepackage{natbib}  
\usepackage{caption} 
\usepackage{algorithm}
\usepackage{algorithmic}
\usepackage{amsthm,amsmath,amssymb}
\usepackage{mathrsfs}
\usepackage{color}
\usepackage{newfloat}
\usepackage{listings}
\usepackage{amsfonts,amssymb}
\usepackage{caption} 
\DeclareCaptionStyle{ruled}{labelfont=normalfont,labelsep=colon,strut=off} 
\floatstyle{ruled}
\newfloat{listing}{tb}{lst}{}
\floatname{listing}{Listing}
\title{
Why mamba is effective? Exploit Linear Transformer-Mamba Network for Multi-Modality Image Fusion}

\author{
   Chenguang Zhu\textsuperscript{\rm 1}, Shan Gao, Huafeng Chen, Guangqian Guo, Chaowei Wang, Yaoxing Wang, Chen ShuLei, QuanjiangFan
}
\affiliations{
   
    Northwestern Polytechnical University\\
    zcg23@mail.nwpu.edu.cn

}

\begin{document}

\maketitle

\begin{abstract}
Multi-modality image fusion aims to integrate the merits of images from different sources and render high-quality fusion images. However, existing feature extraction and fusion methods are either constrained by inherent local reduction bias and static parameters during inference (CNN) or limited by quadratic computational complexity (Transformers), and cannot effectively extract and fuse features. To solve this problem, we propose a dual-branch image fusion network called Tmamba. It consists of linear Transformer and Mamba, which has global modeling capabilities while maintaining linear complexity. Due to the difference between the Transformer and Mamba structures, the features extracted by the two branches carry channel and position information respectively. T-M interaction structure is designed between the two branches, using global learnable parameters and convolutional layers to transfer position and channel information respectively. We further propose cross-modal interaction at the attention level to obtain cross-modal attention. Experiments show that our Tmamba achieves promising results in multiple fusion tasks, including infrared-visible image fusion and medical image fusion.  Code with checkpoints will be available after the peer-review process.
\end{abstract} 

%

\begin{figure}[t]
\centering
\includegraphics[width=1\columnwidth]{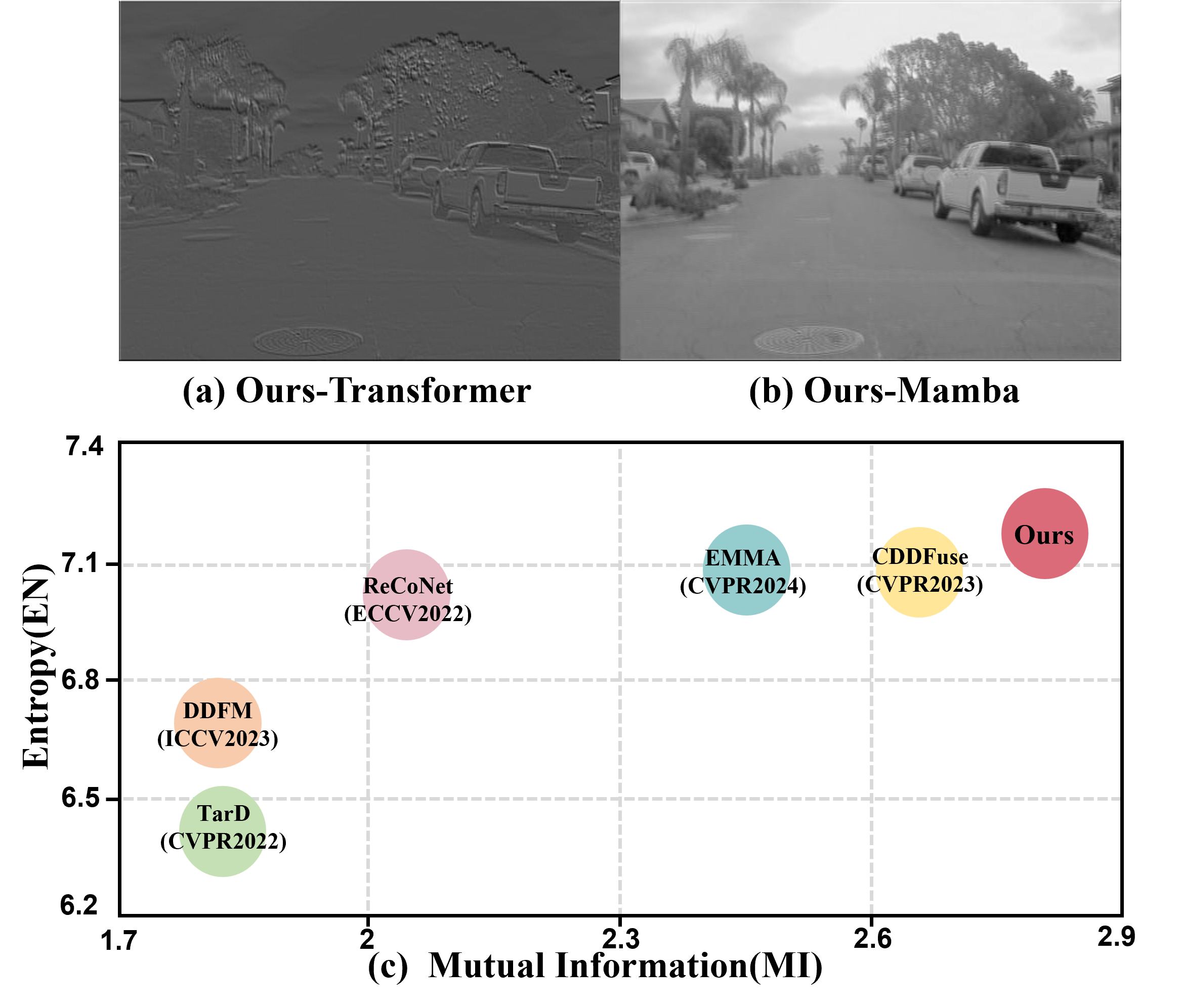}
 \caption{(a) and (b) visualize the features extracted by the Transformer and Mamba branches. It can be seen that the feature patterns of the two branches are very different. (c) Comparison of EN and MI values of fused images by different methods. The fusion results obtained by our method are rich in information and well preserve the input image information.}\label{fig1}
\end{figure}

\section{Introduction} 
Image fusion can integrate important image information from different data sources to render fused images with rich information.
Traditional image fusion approaches typically employ hand-crafted feature extraction and fusion rules. With the advancement of deep learning, image fusion methods based on deep learning have attained remarkable results~\cite{c:31,c:4,c:5,c:6}. In recent years, the networks used for feature extraction and image fusion are mostly built by CNN and Transformer. CNN-based methods have difficulty in capturing global context due to their limited receptive field, which makes them challenging to obtain high-quality fused images~\cite{c:20}. The convolution kernel of CNN has static parameters during the inference phase, which reduces the adaptability of CNN to different inputs \cite{c:34}.
The Transformer-based methods perform well in global modeling, but the quadratic complexity of self-attention leads to high computational overhead. People use CNN and Transformer to form the dual-branch network to achieve the complementarity of network structure, but the problems of low input adaptability of CNN in the inference stage and quadratic complexity of Transformer still exist, which limits its application in the field of image fusion.

Recently, Mamba \cite{c:39} provides us more options. Compared with CNN, Mamba has global modeling capabilities, and its selective scanning mechanism enables it to have high input adaptability. Compared with Transformer, Mamba has linear complexity. However, compared with the dual-branch network composed of CNN and Transformer, the ability of Mamba to extract features is obviously insufficient. The richness of features has a great influence on the effect of image fusion task. Therefore, how to customize a dual-branch network for Mamba suitable for multi modality image fusion is a compelling research issue.

In this paper, we propose the Tmamba block and use it to build a fusion network. It is a dual-branch network tailored for Mamba based on image fusion requirements. We considered three main points when choosing the branch to pair with Mamba: First, Mamba is good at processing long sequence inputs, so the branch paired with it should maintain acceptable computational complexity under long sequence inputs. Second, the branch paired with Mamba needs to be complementary and optimized with Mamba. Third, the branch paired with Mamba cannot undermine Mamba's advantages in image fusion. We choose Restormer block \cite{c:34} and Vmamba block \cite{c:18} to build the dual-branch Tmamba block.  Restormer block is a Transformer with linear complexity, which allows us to use pixel-level sequences while maintaining reasonable computational complexity. The design principles and network structures of Restormer block and Vmamba block are different, which enables them to extract features of different patterns as shown in Figure 1 (a) and (b). Diverse features can provide richer information for fusion. In addition, different network structures also make the features carry different information. Restormer block implicitly implements global attention through channel interactions, so that the features it extracts carry channel attention information. Mamba’s forgetting mechanism gives position information to the features it extracts \cite{c:38}.  Unlike traditional dual-branch networks, two branches in our network are not completely independent. We added an interaction structure between the two branches to help complementary information pass to each other. Both Transformer and Mamba can adjust parameter matrices according to different inputs for targeted inference. This enables our method to fully extract and preserve information from different modalities. As shown in Figure 1 (c), entropy (EN) reflects the amount of information in the fused image, and mutual information (MI) reflects the similarity between the fused image and the input images. Our method can not only extract rich information, but also well preserve the features of the input images.

Furthermore, considering that the original self-attention of a single modality may be highly restricted by the modality information, we designed a cross-modal interaction at the attention level to get cross-modal attention.

In summary, the contributions of our work are as follows:

\begin{itemize}
\item We innovatively built a Transformer-Mamba hybrid framework for multi-modality fusion tasks, and designed a hierarchical interaction strategy between Mamba and Transformer to optimize the features extracted by each other.
\item We proposed a cross-modal interaction at the attention level to break the limit of single modality information on attention, and get a more favorable  attention for image fusion .
\item Our model is highly adaptable to image inputs of multiple modalities and is able to perform targeted inferences based on the input images. This enables us to obtain state-of-the-art results on multiple datasets across two tasks by training on merely one dataset.

\end{itemize}

\section{ Related Work}
\subsection{DeepLearning-based Methods for Image Fusion}
With the advancement of artificial intelligence technology, DL-based methods have become the mainstream in the field of image fusion.  DL-based image fusion methods can be roughly divided into three categories: modals based on generative method \cite{c:31,ma2020infrared,c:3,c:4}, Autoencoder based models \cite{c:5,c:6,c:7,c:8,c:9}  and models that combine downstream tasks \cite{c:4,c:10,c:11}. (1)Among the image fusion methods based on generative methods, GAN-based methods are more commonly used. GAN-based models integrate different visual information by establishing an adversarial game between the original image and the fusion result, generating image results with richer and more diverse content. Recently, \cite{c:31} proposed generative image fusion networks based on diffusion models that leverage powerful generative priors to address challenges such as training instability and lack of interpretability of GAN-based generative methods. (2)Autoencoder-based models extract image features using the encoder, then fuse the extracted image features, and finally output the final image through the decoder. (3)Many works combinine multi-modal image fusion with downstream tasks, \cite{c:11} makes the model obtain more semantic information by introducing segmentation loss,  \cite{c:4} explores methods to combine image fusion with detection, and \cite{c:10} achieves good results in both tasks through the joint optimization of fusion network and segmentation network.

\subsection{State Space Model}
SSM is a fundamental scientific model used in control theory. In recent years, efforts have been made to apply it to deep learning related tasks. LSSL \cite{c:12} is the first to introduce SSM into the field of deep learning, demonstrating its significant advantages in handling long-sequence speech classification tasks. S4 \cite{c:13} introduces low-rank correction adjustment based on LSSL and reduces the computational complexity of SSM. Subsequent works S5 \cite{c:14} and H3 \cite{c:15} improve SSM to make it better adapted to tasks in the field of deep learning. Recently, the emergence of Mamba \cite{c:16} has once again triggered a craze for SSM in the field of deep learning. It can selectively extract input features and also integrate  hardware-aware algorithm. In the computer vision community, many models based on Mamba have also emerged. Vision mamba \cite{c:17} and Vmamba \cite{c:18} provide spatial perception capabilities for Mamba through bidirectional sequence modeling and cross-scanning, respectively. Many mamba-based models \cite{c:19,c:20,c:21} have also emerged and achieved good results In the multi-modality image fusion. They all choose to use Mamba instead of Transformer and CNN, but did not try to combine them.

\begin{figure*}[t]
\centering
\includegraphics[width=1\textwidth]{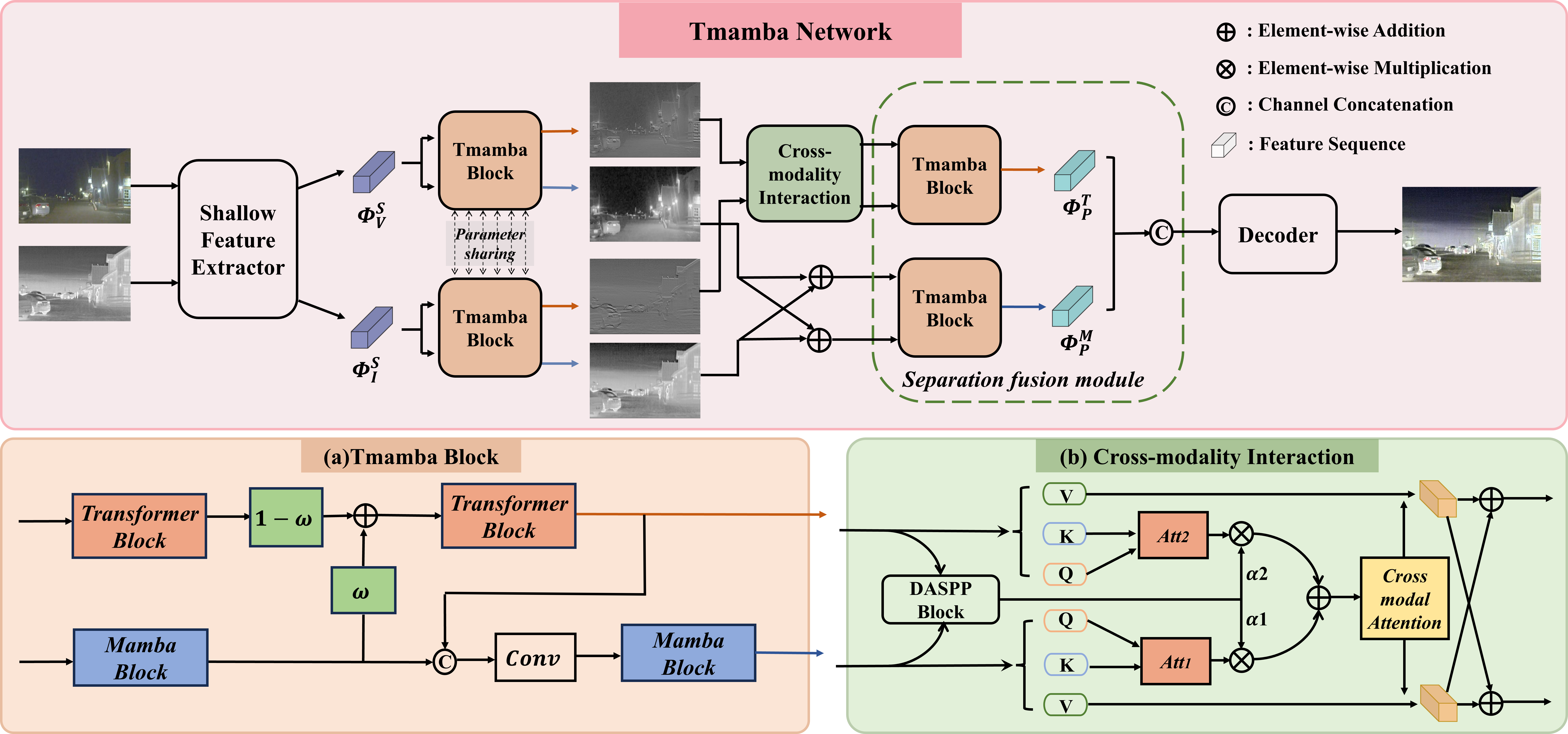} 
\caption{(\textbf{a}) The structure of the Tmamba block. 
(\textbf{b}) The structure of the Cross-modality Interaction module.We use the DASSP block \cite{c:36} to assign optimal weights to the attention of the two modalities.}
\end{figure*}

\section{Method}

\subsection{Overview}

Our network can be specifically divided into five parts. Shallow feature extractor, Tmamba block for further feature encoding and feature interaction, Cross-modal interaction module for achieving cross-modal attention, Tmamba fusion module for fusing images of different patterns separately and Decoder for rendering the final fused image. All Transformer structures in our network are built by Restormer blocks \cite{c:34} and all Mamba structures in our network are built by Vmamba blocks \cite{c:18}. Our network is a generic image fusion network and we will explain our work using  Infrared-Visible Image Fusion as an example.

\subsection{Shallow feature extractor} 
Here we define the input paired infrared and visible images as $I\in\mathbb{R}^{ H\times W}$ and $V\in\mathbb{R}^{H\times W}$.  The shallow feature extractor, Mamba and the Transformer blocks are represented by $\mathcal{S(\cdot)}$, $\mathcal{M(\cdot)}$ and $\mathcal{T(\cdot)}$ respectively.

We first use Transformer blocks to form a shallow feature extractor to extract shallow features \{$\Phi_I^S$, $\Phi_V^S$\} from infrared and visible inputs \{$I , V$\} :

\begin{equation}\label{1}
\Phi_I^S = \mathcal{S}(I) \enspace,\enspace \Phi_V^S = \mathcal{S}(V).
\end{equation}

In order to take advantage of Mamba in processing long sequence features, the shallow feature extractor processes the input image into $128\times128\times64$ features. 

\subsection{Tmamba Block} 

We build Tmamba blocks with two layers of Mamba blocks(Mamba branch) and two layers of Transformer blocks(Transformer branch). We add an interaction structure between the two branches so that the two branches can transfer information to each other while maintaining feature differences.

\subsubsection{Transformer Branch.}  The shallow features extracted by the shallow feature extractor are input into the Transformer branch:

\begin{equation}\label{3}
\Phi_I^{trans},\Phi_V^{trans} = \mathcal{T}(\Phi_I^S,\Phi_V^S),
\end{equation}
where $\Phi_{\{I,V\}}^{trans}$ represent features extracted from the first Transformer block. 

Restormer block can implicitly model global features by self-attention across channels.   $\Phi_{\{I,V\}}^{trans}$ are encoded as $Q\in\mathbb{R}^{ HW\times C}$, $K\in\mathbb{R}^{C\times HW}$, and  $V\in\mathbb{R}^{HW\times C}$. We interact $Q$ and $K$ in the channel dimension to get the channel attention matrix with dimensions $ C \times C $ :

\begin{equation}\label{7}
\mathcal{A} = Softmax(K \times Q  / \alpha),
\end{equation}
where $\mathcal{A}\in\mathbb{R}^{C\times C}$, $\alpha$ is a scaling parameter.
The elements in this matrix reflect the attention relationship between the corresponding channels. After interacting with $V$, this information will be passed to $\Phi_{\{I,V\}}^{trans}$

\subsubsection{Mmaba Branch.} The input features of the Mamba branch are the same as those of the Transformer branch:

\begin{equation}\label{3}
\Phi_I^{vm},\Phi_V^{vm} = \mathcal{M}(\Phi_I^S,\Phi_V^S),
\end{equation}
where $\Phi_{\{I,V\}}^{vm} $ represent features extracted from the first Mamba block respectively.

 Mamba’s forgetting mechanism ensures that the previous hidden state is always attenuated until the current token
is reached, which makes Mamba sensitive to the order of the input sequence and has position awareness \cite{c:38}. This allows each feature pixel in $\Phi_{\{I,V\}}^{vm} $ to sense the positions of other feature pixels.

\subsubsection{T-M Interaction.}


The information extracted by the two branches can complement each other. The channel attention information of the features extracted by the Transformer branch is exactly what the Mamba branch lacks, and the position information of the features extracted by the Mamba branch can also help the Transformer better extract and parse the input image without position encoding.

We designed different interaction structures based on the characteristics of information to help the two types of  information transmit between the Transformer branch and the Mamba branch.
To be specific, we choose a method similar to positional encoding to pass positional information  to the Transformer block. We align and add the features with position information  to the features extracted by the Transformer branch. In the process of addition, we set a global learnable parameter $\omega$, which controls the proportion of information transmitted from the Mamba branch to the Transformer branch without interfering with the position information  encoded by Mamba:

\textbf{\begin{equation}\label{4}
\begin{split}
\Phi_I^{T} = \mathcal{T}(\omega\cdot\Phi_I^{vm}+(1-\omega)\cdot\Phi_I^{trans}),
\\
\Phi_V^{T} = \mathcal{T}(\omega\cdot\Phi_V^{vm}+(1-\omega)\cdot\Phi_V^{trans}) ,
\end{split}
\end{equation}}
where $\Phi_{\{V,I\}}^{T}$ are the final outputs of the Transformer branch

For features with channel attention, we use 1×1 convolution kernel to mix them  with features extracted by Mamba branch, and then use 3×3 convolution kernel to aggregate local features in space and send the aggregated features into Mamba block:

\begin{equation}\label{5}
\begin{split}
\Phi_I^{M} = \mathcal{M}(\mathcal{C}onv(\Phi_I^{vm},\Phi_I^{T})),
\\
\Phi_V^{M} = \mathcal{M}(\mathcal{C}onv(\Phi_V^{vm},\Phi_V^{T})),
\end{split}
\end{equation}
where $Conv(\cdot)$ stands for convolution operation and $\Phi_{\{V,I\}}^{M}$ are the final outputs of the Mamba branch.

The interactive process allows the features of the two branches to be optimized while maintaining their differences, which enables the Tmamba block to extract rich and high-quality image features.
\subsection{Cross-modality Interaction}
After passing through the two branches of the Tmamba block, each modality (infrared and visible) has two different patterns of features due to the different network structures of the two branches.  Before entering the fusion module, we integrate and process the same pattern features of different modalities. For the features extracted from the Mamba branchs, we align and add the features at element wise to obtain the pre-fusion features:

\begin{equation}\label{6}
\Phi_P^{M} = \Phi_I^{M} + \Phi_V^{M},
\end{equation}
where $\Phi_P^{M}$ is the pre-fusion feature of the Mamba branchs.

For the features extracted from the Transformer branchs, we adopt the  cross-modality interaction at attention-level , which simply refers to $\mathcal{AF}$.

The $\mathcal{AF}$ block consists of two attention blocks and a weight calculation block $\mathcal{W}(\cdot)$. Specifically, we first obtain the  attention matrices of the infrared and visible modality through the operation of channel interaction respectively. In the process of calculating attention, the features extracted by the network are encoded as $Q\in\mathbb{R}^{ HW\times C}$,$K\in\mathbb{R}^{C\times HW}$, and  $V\in\mathbb{R}^{HW\times C}$. We multiply K and Q matrix to get  attention matrix with dimensions $ C \times C $.

\begin{equation}\label{7}
\begin{split}
\mathcal{A}_V &= Softmax(K_V \times Q_V  / \alpha),
\\
\mathcal{A}_I &= Softmax(K_I \times Q_I  / \beta) ,
\end{split}
\end{equation}
where $\mathcal{A}_V,\mathcal{A}_I\in\mathbb{R}^{C\times C}$, $\alpha$ and $\beta$ are scaling parameters.

Since there is interaction between channels in the calculation process of the Transformer branch’s attention, the attention matrix reflects the attention of each channel to other channels. Obviously, for different modalities of the same scene, the attention relationship between the various channels is different. We choose the DenseASPP to build  $\mathcal{W}(\cdot)$ to generate weights and get the cross-modal attention. 
Specifically, we first use DenseASPP blocks to further encode the features of the two modalities, and then connect the encoded features in the channel dimension and send them into the full connection layer to obtain two weights:

\begin{equation}\label{7}
\begin{split}
\omega_1 , \omega_2 = \mathcal{W}(\Phi_V^{T},\Phi_I^{T}).
\\
\mathcal{A} = \omega_1  \cdot \mathcal{A}_V + \omega_2 \cdot \mathcal{A}_I ,
\end{split}
\end{equation}

Finally, we apply the cross-modal attention to the V of the two modalities separately, and then add them at element wise to obtain the pre-fusion feature:

\begin{equation}\label{9}
\Phi_P^{T} = \mathcal{A}\cdot V_I + \mathcal{A}\cdot V_V
\end{equation}
where $\Phi_P^{T}$ is the pre-fusion feature of the Transformer branchs, $V_I$ and $V_V$ $\in\mathbb{R}^{HW\times C}$ are the encoded  matrices.


\subsection{Tmamba Fusion}

The two patterns of features extracted by the Transformer and Mamba branches are fused separately in the fusion layer. Considering that the inductive bias of feature fusion should be similar to that of feature extraction, we still adopt Tmamba block for fusion. For the fusion block of Mamba's features, we only select the output of Mamba branch as the final output. and for fusion block of Transformer's features, we only select the output of Transformer branch as the final output:

\begin{equation}\label{10}
\Phi_M = \mathcal{F}_\mathcal{M}(\Phi_P^{M}) \enspace ,\enspace \Phi_T = \mathcal{F}_\mathcal{T}(\Phi_P^{T}),
\end{equation}
where $\mathcal{F}_\mathcal{M}(\cdot)$ and $\mathcal{F}_\mathcal{T}(\cdot)$
 are the fusion blocks of Transformer's features and Transformer's features respectively.

\subsection{Decoder} 
We keep the decoder structure consistent with the design of shallow feature encoder, using the Restormer block as the basic unit of the decoder.

Our training process is divided into two stages. In the first stage, the model does not fuse images, but performs image restoration. In the second stage, we add the Cross-modality Interaction module and Tmamba fusion module to the training to obtain high quality fusion images:
\begin{equation}\label{11}
\begin{split}
&Stage\uppercase\expandafter{\romannumeral1}: V = \mathcal{D}(\Phi_V^{T},\Phi_V^{M}) ,  I = \mathcal{D}(\Phi_I^{T},\Phi_I^{M}).
\\
&Stage\uppercase\expandafter{\romannumeral2}: F = \mathcal{D}(\Phi^{T},\Phi^{M}),
\end{split}
\end{equation}
where $\mathcal{D}(\cdot)$ indicates that the Decoder module, $F$ is the final fusion image.

\subsection{Loss Function}

The loss of $Stage\uppercase\expandafter{\romannumeral1}$ is the image reconstruction loss, which is designed to guide the encoder and decoder to learn the basic feature extraction and image reconstruction methods. The specific loss function is as follows:

\begin{equation}\label{12}
\mathcal{L}_{\uppercase\expandafter{\romannumeral1}} = \mathcal{L}_{ir} + \mathcal{L}_{vis},
\end{equation}
where $\mathcal{L}_{ir}$ and $\mathcal{L}_{vis}$ are reconstruction losses of infrared image and visible image respectively.The reconstruction loss of infrared image can be concretely written as:
\begin{equation}\label{13}
\mathcal{L}_{ir} = \mathcal{L}^{\uppercase\expandafter{\romannumeral1}}_{int}(I,\hat{I}) + \mathcal{L}_{SSIM}(I,\hat{I}),
\end{equation}
where $\mathcal{L}^{\uppercase\expandafter{\romannumeral1}}_{int}(I,\hat{I}) = \|I-\hat{I}\|^{2}_{2}$ and $\mathcal{L}_{SSIM}(I,\hat{I}) = 1-SSIM(I,\hat{I})$,$SSIM(\cdot,\cdot)$is the structural similarity index\cite{c:37}.The reconstruction loss of visible image can also be obtained by the same method.

The loss of $Stage\uppercase\expandafter{\romannumeral2}$ is used to guide the process of image fusion, which mainly consists of intensity loss and gradient loss:
\begin{equation}\label{14}
\mathcal{L}_{\uppercase\expandafter{\romannumeral2}} = \mathcal{L}^{\uppercase\expandafter{\romannumeral2}}_{int} + \mathcal{L}_{grad}
\end{equation}
where $\mathcal{L}^{\uppercase\expandafter{\romannumeral2}}_{int} = \frac{1}{HW}\|I_{f} - max(I_{ir},I_{vis})\|_1$ and $\mathcal{L}_{grad} = \frac{1}{HW} = \||\nabla I_{f}| - max(|\nabla I_{ir}|,|\nabla I_{vis}|)\|_1 $,where $\nabla$ indicates the Sobel gradient operator.

\section{Experiments}

\subsection{Infrared and visible image fusion}
\subsubsection{Setup}
 IVF experiments use three popular benchmarks to verify our fusion model, i.e., TNO \cite{c:24}, RoadScene\cite{c:23}, and  MSRS \cite{c:22}. We train our model on MSRS training set (1083 pairs) and test it on MSRS test set (361 pairs), RoadScene (50 pairs) and TNO (25 pairs). Note that we did not retrain on the TNO and RoadScene datasets, but directly tested on the test sets of the three datasets using the model trained on the MSRS. We use six metrics to quantitatively measure the fusion results: entropy (EN), standard deviation (SD), spatial frequency (SF), mutual information (MI) ,visual information fidelity (VIF) and QAB/F . Higher metrics indicate that a fusion image is better. The details of these metrics can be found in \cite{c:25}.

 Our experiments are carried out on a machine with one NVIDIA GeForce RTX 3090 GPUs. The training samples are randomly cropped into 128×128 patches in the preprocessing stage. The number of epochs for training is set to 80 with 40 and 40 epochs in the first and second stage respectively. The batch size is set to 4. We adopt the Adam optimizer with the initial learning rate set to $7.5 \times 10^{-5}$ and decreasing by 0.5 every 20 epochs. 

\begin{table}[t]
	\centering                                       
        \tabcolsep=0.9mm
	\begin{tabular}{ccccccc}                         
		\hline                                   
		\multicolumn{7}{c}{\textbf{Dataset: TNO Infrared-Visible Fusion Dataset}}\cr
		&EN &SD &SF&MI &VIF &Qbaf\\
		  \hline                                    
		DID\cite{c:26}&6.97&45.12&12.59&1.70&0.60&0.40\\
		U2F\cite{c:27}&6.83&34.55&11.52&1.37&0.58&0.44\\
		TarD\cite{c:4}&6.84&45.63&8.68&1.86&0.53&0.32\\
            ReC\cite{c:29}&7.10&44.85&8.73&1.78&0.57&0.39\\
            CDD\cite{c:9}&\textcolor{blue}{7.12}&46.00&\textcolor{red}{13.15}&\textcolor{blue}{2.19}&\textcolor{blue}{0.77}&\textcolor{blue}{0.54}\\
            DDF\cite{c:31}&6.80&33.15&7.91&1.60&0.64&0.44\\
            EMM\cite{c:32}&7.16&\textcolor{blue}{46.78}&11.74&2.12&0.70&0.49\\
            \textbf{Ours}&\textcolor{red}{7.21}&\textcolor{red}{46.81}&\textcolor{blue}{13.11}&\textcolor{red}{2.36}&\textcolor{red}{0.83}&\textcolor{red}{0.56}\\
		\hline\\[-1.2ex]                                  
  \multicolumn{7}{c}{\textbf{Dataset: RoadScene Infrared-Visible Fusion Dataset}}\cr
		&EN &SD &SF&MI &VIF &Qbaf\\
		  \hline                                    
		DID\cite{c:26}&7.43&51.58&14.66&2.11&0.58&0.48\\
		U2F\cite{c:27}&7.09&38.12&13.25&1.87&0.60&0.51\\
		TarD\cite{c:4}&7.17&47.44&10.83&2.14&0.54&0.40\\
            ReC\cite{c:29}&7.36&52.54&10.78&2.18&0.59&0.43\\
            CDD\cite{c:9}&\textcolor{blue}{7.44}&\textcolor{blue}{54.67}&\textcolor{blue}{16.36}&\textcolor{blue}{2.30}&\textcolor{blue}{0.69}&\textcolor{blue}{0.52}\\
            DDF\cite{c:31}&7.01&36.60&7.25&1.95&0.66&0.47\\
            EMM\cite{c:32}&7.40&53.79&14.37&2.27&0.66&0.47\\
            \textbf{Ours}&\textcolor{red}{7.58}&\textcolor{red}{60.18}&\textcolor{red}{16.79}&\textcolor{red}{2.35}&\textcolor{red}{0.73}&\textcolor{red}{0.52}\\

            \hline\\[-1.2ex]                                   

		\multicolumn{7}{c}{\textbf{Dataset: MSRS Infrared-Visible Fusion Dataset}}\cr
		&EN &SD &SF&MI &VIF &Qbaf\\
		  \hline                                    
		DID\cite{c:26}&4.27&31.49&10.15&1.61&0.31&0.20\\
		U2F\cite{c:27}&5.37&25.52&9.07&1.40&0.54&0.42\\
		TarD\cite{c:4}&5.28&25.22&5.98&1.49&0.42&0.18\\
            ReC\cite{c:29}&6.61&43.24&9.77&2.16&0.71&0.50\\
            CDD\cite{c:9}&6.70&\textcolor{blue}{43.38}&11.56&\textcolor{blue}{3.47}&\textcolor{blue}{1.05}&\textcolor{blue}{0.69}\\
            DDF\cite{c:31}&6.19&29.25&7.46&1.88&0.74&0.48\\
             EMM \cite{c:32}&\textcolor{blue}{6.71}&\textcolor{red}{44.13}&\textcolor{blue}{11.56}&2.94&0.97&0.64\\
            \textbf{Ours}&\textcolor{red}{6.72}&43.32&\textcolor{red}{11.57}&\textcolor{red}{3.69}&\textcolor{red}{1.07}&\textcolor{red}{0.71}\\
		\hline                                  
  
	\end{tabular}
        \caption{Quantitative results of the VIF task. \textcolor{red}{Red} and \textcolor{blue}{blue} fonts indicate best and second-best values} 
\end{table}

\subsubsection{Comparison with advanced methods.}

In this section, we test Tmamba on the three test sets and compare the fusion results
\begin{figure*}[t]
\centering
\includegraphics[width=1\textwidth]{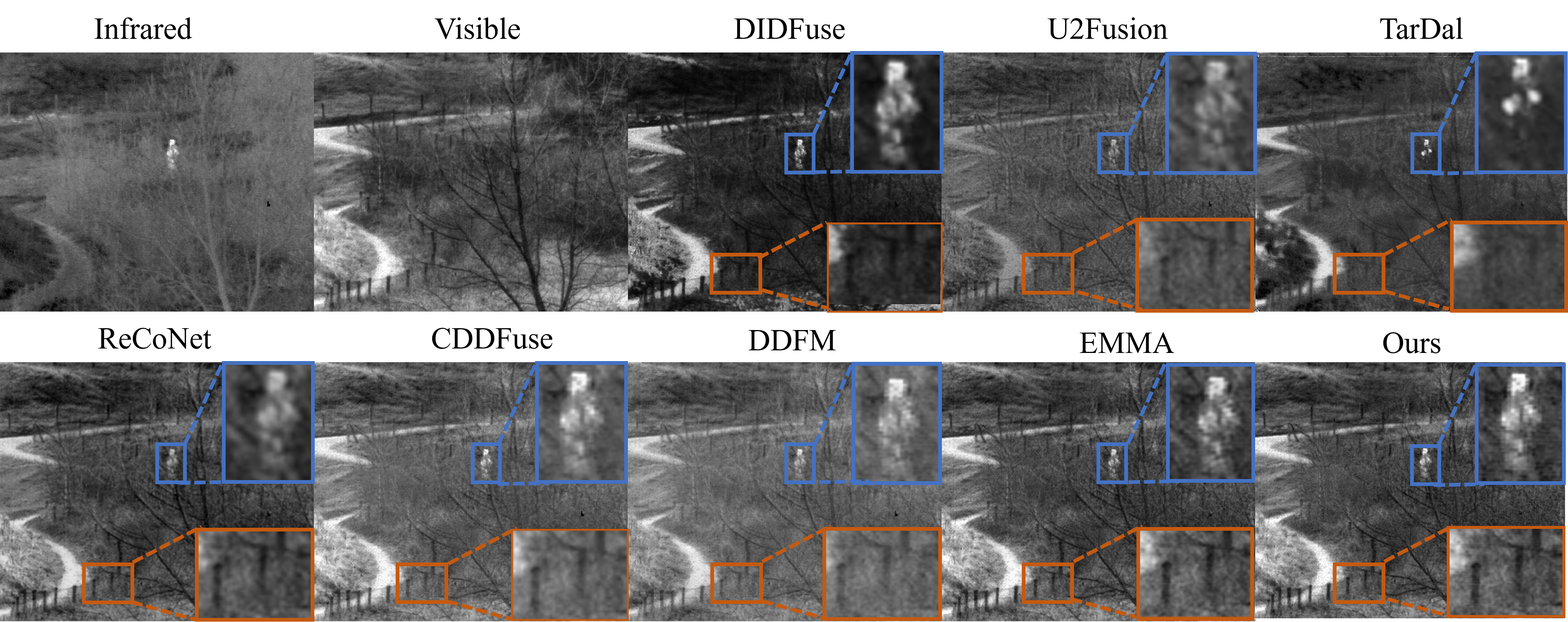} 
\caption{Visual comparison for “18” in sandpath of TNO IVF dataset.}
\end{figure*}

\begin{figure*}[t]
\centering
\includegraphics[width=1\textwidth]{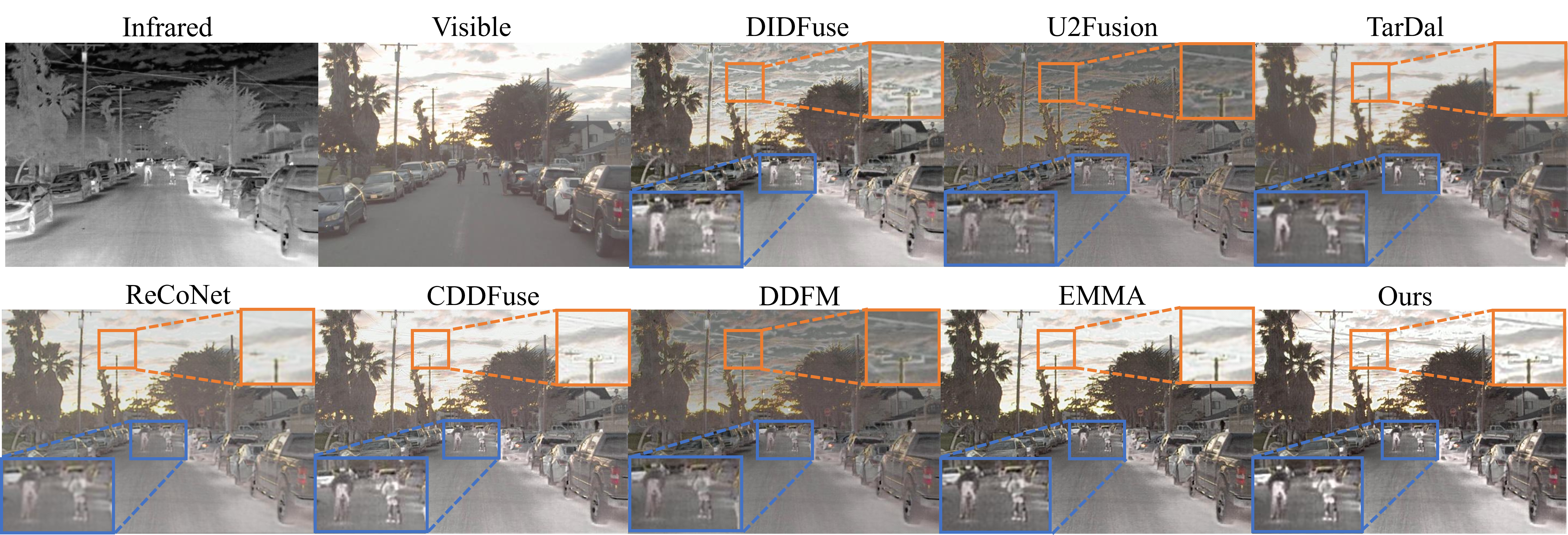} 
\caption{Visual comparison for “FLIR 06506” in RoadScene IVF dataset.}
\end{figure*}

\noindent the state-of-the-art methods including DIDFuse\cite{c:26}, U2Fusion\cite{c:27}, TarDAL\cite{c:4}, ReCoNet \cite{c:29},CDDFuse\cite{c:9},DDFM\cite{c:31},EMMA\cite{c:32}.

For qualitative comparison, We show the qualitative comparison in Figure 3 and Figure 4. Compared with other existing methods, our Tmamba has two significant advantages. First, the discriminative target from infrared images can be well preserved.The people in Figures 3 and Figure 4 are all highlighted, and our method has a higher contrast and clearer outline compared to other methods. Second, our method can get richer background information.As shown in Figure 4 (the wires and poles in the orange box), our method clearly restores the outline of the wires and poles, while visual inspection shows that other methods do not bring out this background information.

For quantitative comparisons, a total of six metrics are employed to quantitatively compare the above results, which are displayed in Table 1. Our method has excellent performance on almost all metrics.

\subsection{Ablation studies}
Ablation experiments are set to verify the rationality of the different modules.We did a total of five experiments to test our model as shown in Table 2 and Table 3.

\subsubsection{Transformer branch.} We remove the Mamba branch from the Tmamba block, leaving only the Transformer branch, while removing the cross-modality interaction of the Transformer branch. The test results show that all the indexes declined. 

\subsubsection{Cross-modality interaction.} We add the cross-modality interaction in the Transformer branch.It makes the attention relationship of each channel no longer restricted by single-modal information, which comprehensively improves the effect of image fusion.

\subsubsection{Mamba branch.} We add the Mamba branch to form a dual-branch structure with Transformer. The test results show that all the indexes improve.  The dual-branch structure composed of Transformer and Mamba can encode image features in different patterns and then fuse them separately, so that the acquired image information is richer and the quality of the fused image is greatly improved.

\subsubsection{Interaction structure.}
We added hierarchical interaction structure between Transformer branch and Mamba branch to form a complete Tmamba block. By adding  hierarchical interaction structure, the  information carried by the the features  is continuously exchanged between the two branches, further improving the fusion effect.

\subsubsection{Why mamba is effective.}  We replace the Mamba block with a residual convolution module. The SF index of the fused image has improved, but the overall quality has decreased, as shown in Table 3. We believe that there are two main reasons. On the one hand, in the long sequence image fusion task, Mamba can show feature extraction and feature analysis capabilities that surpass CNN; on the other hand, we use Transformer and Mamba to build the network backbone, which makes our network input-aware and can adjust parameters for different inputs, making it more suitable for multimodal image fusion tasks than the Transformer-CNN hybrid network.

\begin{figure}[t]
\includegraphics[width=1\columnwidth]{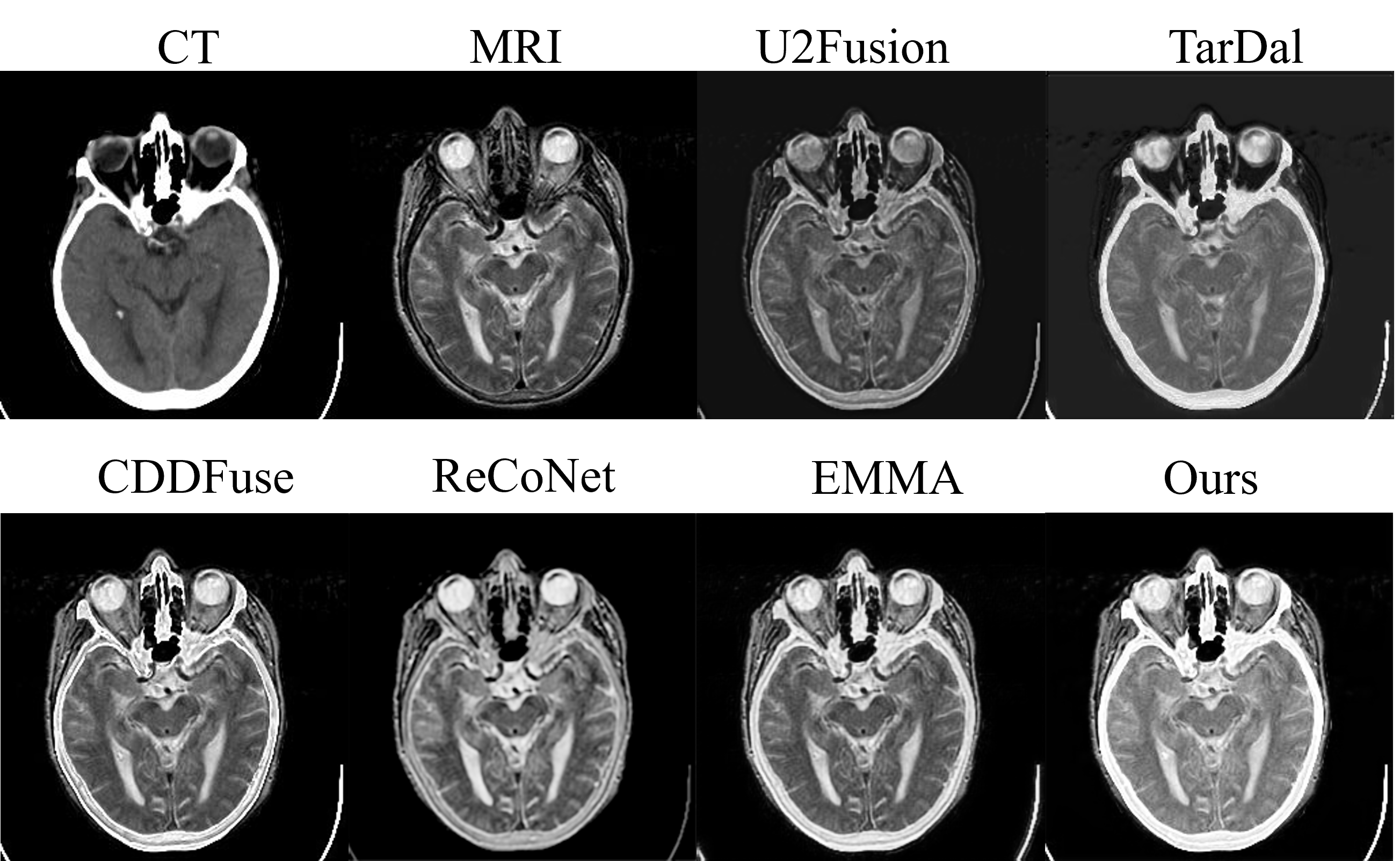}
\caption{Visual comparison for “MRI-CT 14” in  MRI-CT MIF dataset. }

\end{figure}

\begin{table}[t]
	\centering                                       
        \tabcolsep=1.6mm
	\begin{tabular}{cccc|cccccc}                         
		\hline                                   
		\multicolumn{10}{c}{\textbf{Dataset: RoadScene Infrared-Visible Fusion Dataset}}\cr
		\hline  \textbf{T}&\textbf{   A}&\textbf{M}&\textbf{I}&EN&SD&SF&MI&VIF&Qbaf\\
		  \hline                                    
		\checkmark& & & & 7.41&51.55&15.21&2.12&0.66&0.51\\
            \checkmark&\checkmark & & &7.45&54.25&15.15&2.25&0.69&0.52\\
    		\checkmark&\checkmark &\checkmark & & \textcolor{blue}{7.50}&\textcolor{blue}{57.02}&\textcolor{blue}{16.73}&\textcolor{blue}{2.34}&\textcolor{blue}{0.72}&\textcolor{blue}{0.52}\\
		\checkmark&\checkmark &\checkmark & \checkmark&\textcolor{red}{7.58}&\textcolor{red}{60.18}&\textcolor{red}{16.79}&\textcolor{red}{2.35}&\textcolor{red}{0.73}&\textcolor{red}{0.52}\\
           
		\hline                                  
	\end{tabular}
        \caption{Ablation experiment results on the testset of RoadScene. \textcolor{red}{Red} and \textcolor{blue}{blue} fonts indicate the best and second-best values. \textbf{T, A, M}, and\textbf{ I }stand for Transformer branch, Cross-modal attention, Mamba branch, and interaction structure respectively.}
\end{table}

\begin{table}[t]
	\centering                                       
        \tabcolsep=2.1mm
	\begin{tabular}{ccc|cccccc}                         
		\hline                                   
		\multicolumn{9}{c}{\textbf{Dataset: RoadScene Infrared-Visible Fusion Dataset}}\cr
		\hline  \textbf{T}&\textbf{M}&\textbf{C}&EN&SD&SF&MI&VIF&Qbaf\\
		  \hline                                    
		\checkmark& &\checkmark&  7.49&56.00&\textcolor{red}{16.87}&2.33&0.71&0.52\\
            
		\checkmark&\checkmark & &\textcolor{red}{7.58}&\textcolor{red}{60.18}&16.79&\textcolor{red}{2.35}&\textcolor{red}{0.73}&\textcolor{red}{0.52}\\
           
		\hline                                  
	\end{tabular}
        \caption{Ablation experiment results on the testset of RoadScene. Replace the Mamba blocks with CNN blocks.\textcolor{red}{Red}  fonts indicate best  values}
\end{table}

\subsection{Medical image fusion}

\subsubsection{Setup.} We selected 136 pairs of medical images from the Harvard Medical website for MIF experiments, of which 21 pairs of MRI-CT images, 42 pairs of MRIPET images and 73 pairs of MRI-SPECT images are utilized as the test datasets.

\begin{table}[t]
        
	\centering                                       
        \tabcolsep=0.9mm
	\begin{tabular}{ccccccc}                         
		\hline                                   
		\multicolumn{7}{c}{\textbf{Dataset: MRI-CT Medical Image Fusion}}\cr
		&EN &SD &SF&MI &VIF &Qbaf\\
		  \hline                                    
            U2F\cite{c:27}&4.71&46.66&19.82&2.05&0.36&0.40\\
		TarD\cite{c:4}&4.75&61.14&28.38&1.94&0.32&0.35\\
            CDD\cite{c:9}&4.83&\textcolor{blue}{88.59}&\textcolor{red}{33.83}&2.24&\textcolor{blue}{0.50}&\textcolor{red}{0.59}\\
             ReC\cite{c:29}&4.41&66.96&20.16&2.03&0.40&0.42\\
             EMM\cite{c:32}&\textcolor{red}{5.40}&76.49&25.64&\textcolor{red}{2.29}&0.49&0.55\\
            \textbf{Ours}&\textcolor{blue}{5.27}&\textcolor{red}{89.35}&\textcolor{blue}{31.67}&\textcolor{blue}{2.28}&\textcolor{red}{0.50}&\textcolor{blue}{0.58}\\
		\hline   \\[-1.2ex]                                 
  \multicolumn{7}{c}{\textbf{Dataset: MRI-PET Medical Image Fusion}}\cr
		&EN &SD &SF&MI &VIF &Qbaf\\
		  \hline                                    
            U2F\cite{c:27}&4.10&41.41&16.36&1.61&0.43&0.39\\
		TarD\cite{c:4}&3.81&57.65&23.65&1.36&0.57&0.58\\
            CDD\cite{c:9}&4.24&\textcolor{blue}{81.72}&\textcolor{red}{28.04}&\textcolor{blue}{1.87}&\textcolor{blue}{0.66}&\textcolor{blue}{0.65}\\
             ReC\cite{c:29}&3.66&65.25&21.72&1.51&0.44&0.51\\
            EMM\cite{c:32}&\textcolor{blue}{4.72}&73.32&25.65&1.79&0.57&0.64\\
            \textbf{Ours}&\textcolor{red}{4.82}&\textcolor{red}{83.89}&\textcolor{blue}{26.46}&\textcolor{red}{1.93}&\textcolor{red}{0.68}&\textcolor{red}{0.65}\\
		\hline   \\[-1.2ex]                                 
  \multicolumn{7}{c}{\textbf{Dataset: MRI-SPECT Medical Image Fusion}}\cr
		&EN &SD &SF&MI &VIF &Qbaf\\
		  \hline                                    
            U2F\cite{c:27}&3.67&36.89&12.74&1.72&0.48&0.46\\
		TarD\cite{c:4}&3.66&53.46&18.50&1.44&0.64&0.52\\
            CDD\cite{c:9}&3.91&\textcolor{blue}{71.82}&\textcolor{red}{20.68}&1.89&\textcolor{blue}{0.66}&0.69\\
             ReC\cite{c:29}&3.22&60.07&17.40&1.50&0.46&0.54\\
             EMM\cite{c:32}&\textcolor{blue}{4.40}&64.17&\textcolor{blue}{20.59}&\textcolor{blue}{1.93}&0.64&\textcolor{blue}{0.69}\\
            \textbf{Ours}&\textcolor{red}{4.42}&\textcolor{red}{72.24}&19.81&\textcolor{red}{1.96}&\textcolor{red}{0.67}&\textcolor{red}{0.69}\\
		\hline                                  
	\end{tabular}
 \caption{Quantitative results of the MIF task.\textcolor{red}{Red} and \textcolor{blue}{blue} fonts indicate best and second-best values}
\end{table}

\subsubsection{Comparison with advanced methods.}  In this group of experiments, we selected five networks, U2Fusion \cite{c:27} TarDAL\cite{c:4}, ReCoNet \cite{c:29}, CDDFuse\cite{c:9}, and EMMA\cite{c:32}, to conduct comparative tests.We made quantitative and qualitative comparisons across all networks. Note that none of the models were fine-tuned on the medical data set. 

For qualitative comparison,we visualize the fusion results on the  MRI-CT dataset in Figure 5. The results of quantitative comparison are shown in Table.4.  As you can see, our method achieves leading performance on all datasets.

\section{Conclusion}

In this paper, we proposed a dual-branch Transformer-Mamba network for multi-modality image fusion. By combining the Restormer block and the Vammba block, we extracted features containing different information. We interacted and fused these features and achieved better results. We further proposed cross-modality Interaction at the attention level to break the limitation of single-modality information on attention. Experiments show that our method achieves advanced performance on six datasets across two image fusion tasks.

\bibliography{aaai25}

\end{document}